%% file: DataMinimalism.tex
\documentclass[runningheads]{llncs}
\usepackage{graphicx}
\usepackage{balance}  
\usepackage{color}
\usepackage{booktabs}
\usepackage{floatrow}
\newfloatcommand{capbtabbox}{table}[][\FBwidth]

\usepackage{hyphenat}
\usepackage[official]{eurosym}

\begin{document}
\title{Computing the Value of Data: Towards Applied Data Minimalism}


\author{Michaela Regneri\inst{1} \and
Julia S. Georgi\inst{1}\and
Jurij Kost\inst{1} \and
Niklas Pietsch\inst{2} \and
\\Sabine Stamm \inst{2}
}

\authorrunning{M. Regneri et al.}
%
\institute{OTTO (GmbH \& Co. KG), Werner-Otto-Str. 1-7, 22179 Hamburg, Germany \email{\{Michaela.Regneri,JuliaSoraya.Georgi,Jurij.Kost\}@otto.de}
\url{https://www.otto.de/unternehmen/en} \and
Otto Group Solution Provider (OSP) GmbH, Hartzloh 25, 22309 Hamburg, Germany 
 \email{\{niklas.pietsch,sabine.stamm\}@ottogroup.com} \\
\url{https://www.osp.de/en} }

\maketitle


 
\input abstract

\input introduction

\input related

\input datavalue

\input experiments
\input conclusion

\bibliographystyle{splncs04}
\bibliography{DataMinimalism}

\end{document}

%% file: abstract.tex
\begin{abstract}
We present an approach to compute the monetary value of individual data points, in context of an automated decision system. The proposed method enables us to explore and implement a paradigm of data minimalism for large-scale machine learning systems. Data minimalistic implementations enhance scalability, while maintaining or even optimizing a system's performance. Using two types of recommender systems, we first demonstrate how much data is ineffective in both settings. We then present a general account of computing data value via sensitivity analysis, and how, in theory, individual data points can be priced according to their informational contribution to automated decisions. We further exemplify this method to lab-scale recommender systems and outline further steps towards commercial data\hyp{}minimalistic applications.
\end{abstract}

%% file: introduction.tex
\section{Introduction}
Nowadays, data is colloquially considered ``the new oil'' as well as ``the new gold''. 
Most data-driven systems, particularly in the field of artificial intelligence (AI), treat data proverbially as ``the new bacon'': They use as much input data as possible, assuming that more information always produces better automated decisions. An abundance of data is mostly seen as an optimal basis for machine learning and data is, if available, treated as a free commodity.

In practice, data sparsity is far more often encountered than the contrary, and a plethora of research has developed systems that robustly accommodate the undesired lack of data. So far, there are hardly any strategies of intentionally cutting input data sources for optimization purposes.
We introduce a different principle to machine learning, promoting \emph{data minimalism} as a paradigm that is beneficial in many aspects:

\begin{enumerate}

\item \textbf{Cost Efficiency:} While data \emph{storage} is comparably cheap, computational costs increase heavily with data \emph{usage}. CPU hours bear either direct costs (e.g. cloud pricing) or indirect costs (like energy and time). Both can be drastically reduced by downsizing the input dataset.

\item \textbf{Societal Responsibility:} Using data as a free commodity has two sides: while the financial aspect might be negligible in some context, there is an increasingly recognized environmental and social dimension. Each byte of data transferred unnecessarily creates safety and privacy risks. Further, each CPU hour caused by redundant data processing causes unnecessary energy use and thus CO$_2$ emissions. Recent work has shown that greenhouse emissions should be particularly concerning for AI research \cite{Co2}. Given the general growth of data worldwide, a paradigm of data leanness is overdue.

\item \textbf{Quality Gains:} Lack of data is often equated with low system performance. We will show that this is not necessarily true: data can carry toxic information that even corrupts a system's performance. It then requires more beneficial data to even out the negative influence. A core purpose of data minimalism consists of eliminating data that corrupts output quality.

\item \textbf{Stability:} Especially in machine learning, using large amounts of data requires sophisticated sampling techniques. While those algorithms often show good and stable performance, their actual output can differ vastly even with unchanged input data, because random sampling means choosing between (often many) equally good options. In some contexts, however, stability is extremely desirable, e.g. to comply with legal requirements. Using fewer data means less sampling is needed and more stability can be achieved.
\end{enumerate}

Our main contributions are threefold.
First, we show how to \emph{quantify redundant data} in large-scale data-driven systems using two examples of recommender systems. 
Second and most importantly, we propose how to compute the \emph{value of individual data points} in order to identify data that can or should be omitted for maximal system performance with minimal data volume. Finally, this research lays the foundation to implement \emph{data minimalism} as a general digital principle.

The paper is structured as follows: After an account of related work (Section~\ref{related}), we outline our take on data minimalism using two variants of recommender systems (Section~\ref{dataminimalism}). We show how to compute the value of individual data points in the context of any system making automated decisions and apply this concept to lab-scale recommender systems (Section~\ref{experiments}). We conclude with a summary of the results and an outlook on future research directions (Section~\ref{conclusion}).

%% file: related.tex
\section{Related Work}\label{related}
Data minimalism is often a necessary evil rather than a desired feature. Data sparsity has induced a wealth of work in different fields (see e.g. \cite{Peng01thesparse,10.1093/oxfordjournals.aje.a010240} among many). Related to minimalism and machine learning, there is a growing surge to build sparser \emph{systems}, which aim to reduce the complexity of, e.g., a neural net in favor of explainability and generalisability \cite{liu2015sparse,SURESH20101149}. Sparse \emph{feature} learning serves a similar purpose and helps to prevent systems from overspecialization \cite{pmlr-v44-atzmon2015,Guo:2013:PMC:2540128.2540326}. The growing field of sparse (or compressed) sensing systematizes such approaches in signal processing \cite{eldar_kutyniok_2012,DBLP:phd/basesearch/Chepuri16}. All of those approaches are reminiscent of the ideas outlined in \cite{Halevy:2009:UED:1525642.1525689}, who claim that due to scalability reasons, large amounts of data with naive learning systems are always preferable over sparse input data with added knowledge engineering. 

Our work follows similar ideas and objectives. With data minimalism as a paradigm, we also want to increase stability, simplicity and explainability of systems, with the latter being also a prerequisite to achieve data minimalism in the first place. In contrast to previous work, our approach manipulates the system's input directly rather than engineering the learning process or the learned structure. While this is in line with approaches that prune learning systems, we will show that input data can, in fact, be reduced drastically if it is done sensibly. 

We consider the value of data as a necessary ingredient to build data minimalistic systems. Computing this value is mostly a business problem, researched within the disciplines of \emph{infonomics} \cite{laney2017infonomics} or \emph{infometrics}. To the best of our knowledge, all previous approaches determine the value of whole data sources according to data-specific \cite{DBLP:journals/corr/abs-1811-04665} or business-specific aspects.
We promote a more precise account of data value. Pricing a whole data source is too coarse-grained and misses the chance to save time and resources spent on a large part of uninformative data. This idea builds on our previous research on sensitivity of machine learning systems \cite{DBLP:journals/corr/abs-1807-07404}, in which we dissected the influence of single data points on a system's output. In our current work, we also use a vector-based recommender along with a more naive one and relate the qualitative output changes we examined earlier to monetary value of input data.

%% file: datavalue.tex
\section{Data Minimalism: \textbf{How much data} do we need?}\label{dataminimalism}
In this section, we explain why and how we challenge the assumption that more data always yields better automated decisions. We first explain some background on recommender systems, which serve as our example application (Section~\ref{recos}). A first experiment quantifies uninformative data in a deterministic system based on association analysis (Section~\ref{datavalue:det_sys_association_analysis}). A second experiment outlines the effects of input data augmentations along the learning curve in a machine learning system (Section~\ref{w2v_intro}).

\subsection{Recommender Systems}\label{recos}
Recommender systems \cite{Resnick:1997:RS:245108.245121} use automated means to help users make decisions with insufficient information. In general, such systems either use collaborative filtering, content-based recommendations or hybrid approaches. Collaborative filtering means that the (implicit or explicit) recommendations of all users are aggregated to find the best recommendations for the current user. Content-based recommendations refer to products previously bought or viewed by the current user and generate recommendations based on similarities with those products. Such similarities can be visual, textual or computed from other meta data (such as prices or brands).

Like many modern recommenders, the systems we consider are hybrid approaches. Given a \emph{seed product} (the product currently viewed by the user), we compute a sorted list of recommendations that are possible \emph{alternatives} for the seed product (\emph{``You might also like...''}). 
Such item-to-item recommendations are widely used in e-commerce to increase conversion rates by helping customers to find the right product and by generally keeping them engaged on the site.

\subsection*{Experimental Recommenders}
Our training data consists of anonymous user interactions with product pages. We define the chronologically sorted series of clicks by one user with breaks shorter than a certain time as a \emph{user session}. While a user session contains events of various types, we only consider clicks on product pages, so the \emph{session length} in our case is the number of product page clicks. 
Based on this data, we build two different types of recommender systems.

First, we use a classic deterministic recommender system based on \emph{association analysis} (similar to \cite{OSADCHIY2019535}). The basic idea of the algorithm is to recommend a product $p_1$ as alternative for a product $p_0$ if many different users viewed $p_1$ and $p_0$ together within the same session. The actual rankings of recommendations are determined by the decreasing order of product co-occurrence frequencies, without any further processing. We call this recommender ``co-occurrence recommender'' (COR).

The second recommender system we analyze is a \emph{machine learning system} using word2vec \cite{DBLP:journals/corr/abs-1301-3781}. In principle, this approach follows the same paradigm as the previous one, in that it considers two products as similar if they were frequently visited by the same user in the same session. We use word2vec because it is a common approach for recommender systems and enables the display of implicit product relations (two products can be similar even if they were never directly visited after one another, but have a transitive similarity). We will refer to this recommender system as ``vector recommender'' (VR).

\subsection*{Conversion Rate as Performance Measure}
\label{calc_conversion_rate}
While the systems we analyze are solely built for experimental purposes, both input data and performance measures base upon user interactions in a productive system. We quantify the economic performance of a recommender system with a hypothetic conversion rate as follows:

\begin{enumerate}

\item For each seed product we compute the top five alternative products.
\item For a chosen time frame of three months we calculate for each pair of a seed product and an alternative product
\begin{itemize}
\item the number of user sessions $n_{views}$, where the alternative product is recommended for the seed product,
\item the number of user sessions $n_{ordered}$, where the alternative is recommended and ordered.
\end{itemize}
\item We estimate the hypothetic conversion rate $CR_{hyp}$ as
\begin{equation}
CR_{hyp} = \frac{\sum_{\forall p}\sum_{\forall r_p} n_{ordered}(p,r_p)}{\sum_{\forall p}\sum_{\forall r_p} n_{views}(p,r_p)}*c
\end{equation}
where $p$ is a seed product, $r_p$ is a top recommendation for $p$, and $c$ is a correction constant derived from the input data and the use case at hand to scale the conversion rate\footnote{The factor $c$ is not considered for the analyses in this paper since we are working with relative changes henceforth.}.
\end{enumerate}
To interpret the results, it is helpful to know that the recommendations from the live system used for reference are influenced by business rules preferring similar products instead of complementary products.

\subsection{Data without any Influence: Co-Occurrence Recommender}
\label{datavalue:det_sys_association_analysis}
Our product recommendations are based on information gained from user sessions and, thus, can change with each website visit. Consequently, productive systems are re-trained regularly with a constant volume of updated interaction data. The first experiment aims to examine the impact of a selected user session on product recommendations under the assumption that a session has a certain $activity$ $time$ frame, i.e. the time span (in days) the respective session is used to calculate recommendations with.

One of the hypotheses is that the impact of a session changes as it proceeds through its activity time frame. A second hypothesis concerns the so-called product categories of a user session: each product is part of a type hierarchy, with top-level labels like ``fashion'' and fine-grained labels like ``running shoes''. We examine the influence of category hetereogenity within a user session on the session's influence. (See \cite{DBLP:journals/corr/abs-1807-07404} for a more detailed description on the product categories we examined.)

As a data basis, we select all user sessions of a predefined time frame. We create a $n \times n$ co-occurrence matrix for all $n$ products occurring in those user sessions. A cell $(x,y)$ in the matrix consists of the absolute co-occurrence count of product $x$ and product $y$, i.e. the frequency of the two products occurring in the same user session. Repeated clicks on the same product within one session are only counted once.

The overall set of recommendations we consider are the top~5 recommendations ${r_{1,p},...,r_{5,p}}$ for each product $p$. (We choose five because in our specific use case, five recommendations are typically visible without scrolling.)

In our experimental setup, we constrain the activity time frame to 51 days. This means that product recommendations given today are calculated with user sessions of the preceding 51~days. We treat this time frame as a rolling window shifting daily over the course of the next 50 days, thus, we cover the full activity time frame of all sessions beginning with day 1. To analyze the impact of those sessions, we compute the product recommendations for each of the 50 consecutive days in their life cycle.


Since the product recommendations are based on the frequency counts of the co-occurrence matrix, only sessions containing a seed product and one of the top recommendation products can influence the recommendations of the respective seed product. For this experiment, we consider the top~5 recommendations 
and define the \emph{contribution to visibility} (CV) score as a measure of how many pairs of seed products and recommendations are in a session, examining all seed products viewed in the respective time frame. 

For 791,571 consecutive user sessions, we calculate the CV score over the succeeding 50 time frames. Consequently, we obtain 50 CV scores per session and evaluate the evolving progress of the scores with a linear regression analysis. Using the slope and intercept, we classify the session's impact as ``no impact" (slope and intercept $= 0$), ``stable" (slope $=  0$, intercept $> 0$), ``decreasing" (slope $< 0$) or ``increasing" (slope $> 0$). Table~\ref{COR:impact} summarizes the distribution of sessions in each class.
\begin{figure}\CenterFloatBoxes 
\begin{floatrow}

\killfloatstyle\capbtabbox{%
\begin{tabular}{lrr}
\toprule
impact & \# sessions & percentage \\
\midrule
no impact & 179,446 & 22.7 \\
stable & 266,633 & 33.7 \\
increasing & 166,553 & 21.0 \\
decreasing & 178,939 & 22.6 \\
\bottomrule
\end{tabular}

}
{%
\caption{COR: Overview of user session impact\label{COR:impact} }
}

\ffigbox{%
\includegraphics[scale=0.4]{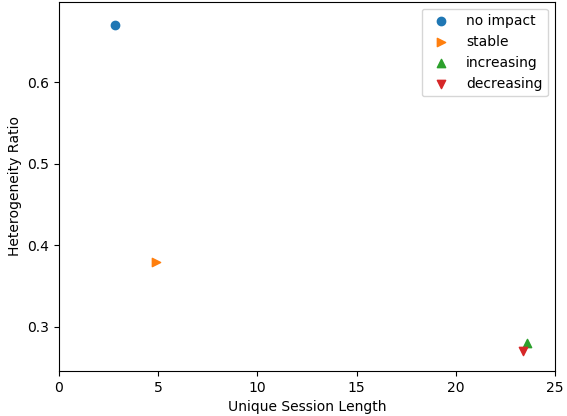}\caption{Mean session length vs. mean heterogeneity ratio for groups from table~\ref{COR:impact}}
\setcounter{figure}{0}

}{%
  \caption{Session length vs. Heterogeneity\label{COR:het_len}}%
}
\end{floatrow}
\end{figure}


22.7\% of the user sessions have no influence on the product recommendations over all 50 time frames, which means the recommendations given by the co-occurrence recommender would not change if those 22.7\% user sessions were left out. The remaining 77.3\% of the sessions influence the product recommendations with varying impact: while 33.7\% of the sessions are stable, the impact of 21\% of the sessions increases and 22.6\% loose influence throughout the activity time frame.

As one indicator for informational contribution, we define the \emph{heterogeneity ratio (HR)} of a user session as 
\begin{equation}
HR = \frac{\textrm{\# of unique product categories}}{\textrm{\# of unique products}}
\end{equation}
Looking at the mean ratio as well as at the mean of the unique session lengths, the differences between the groups are highly significant ($p<0.001$). Figure~\ref{COR:het_len} shows the mean values of the heterogeneity ratio and the unique session length of the four groups from Table~\ref{COR:impact}. 


On average, user sessions without impact on product recommendations tend to have only a small number of different products (measured as small unique session length) but a high heterogeneity ratio, meaning that the few products within those sessions do not share the same product category.
Stable sessions tend to have more products of the same category than those without impact and show a smaller heterogeneity ratio. User sessions with increasing or decreasing influence on average contain many more different products while being comparatively homogeneous.

\subsection{Effects on the Learning Curve: Vector Recommender}
\label{w2v_intro}
In the following, we show how a machine learning recommender system changes with increasing data volume. As Key Performance Indicators (KPIs) we outline the computing time (which can be translated to computing costs) and the algorithm's output quality (measured as revenue), along with other informative figures (number of products in the model and conversion rate).

The vector-based recommendations are computed using word2vec, which is originally designed to compute word embeddings. Words are represented by vectors, and the computed embeddings represent the words in their predicted contexts. In contrast to count-based co-occurrence models, word2vec uses a neural network to compute the vector space, deriving abstract concepts as dimensions for each word from a given corpus. Count-based models define the vector space by directly using context words as dimensions, resulting in many more dimensions that are potentially less effective. Neural models perform superior in numerous tasks \cite{marcobaroni2014predict}.

Technically, seed product and recommendations in our model share similar click embeddings, i.e. other customers looked at the alternatives in similar user journey contexts in which they visited the current product in focus.
Operating on user sessions, we can transfer linguistic terminology: one click in a session corresponds to a \emph{token}, and one abstract product relates to a \emph{(product) type}. In this context we could use the term \emph{product} interchangeably with \emph{type}. The vector model we compute contains one vector per product type.

All experiments are performed on a machine with two Intel(R) Xeon(R) Gold 6126, with a total of 24 physical cores. We compute the product embeddings using Google's word2vec WORD VECTOR estimation toolkit, version 0.1c\footnote{Hyper-parameters: skip-gram (-cbow 0), hierarchical softmax (-hs 1), no sub sampling (-sample 0), five iterations (-iter 5), 200 dimensions (-size 200), 8 threads (-threads 8) and ignore products that occur less than 5 times (-min-count 5).}. We fix the context window size to exactly 5, round all vector numbers to 4 rather than 8 digits and sort the vocabulary (see also~\cite{DBLP:journals/corr/abs-1807-07404}). The consumption of CPU hours are measured for 2, 10, 30, 60, 90, 120, 150, 180, 210, 240, 270 and 300 consecutive days of user sessions.
The considered sessions have been selected backwards from a starting day, such that all models share the same most recent user session, and added data are historical sessions.
\begin{table}
\centering
\small
\begin{tabular}{rrrrcp{2mm}r}

\toprule
\#Days & \#Sessions & \#Products & \multicolumn{1}{c}{SNP(\%)} &  Length& & CPU sec. \\
\midrule
2 & 1,148,450 & 219,787 & 100.00  & 6.94 &	&	1,219.21	\\
10 & 5,493,601 & 496,988 & 28.76  & 7.04 &	&	7,504.02	\\
30 & 16,162,834 & 728,171 & 13.28  & 6.96 &	&	23,710.45	\\
60 & 32,415,119 & 863,687 & 5.61  & 6.99 &	&	44,279.31	\\
90 & 48,087,277 & 939,375 & 2.31  & 6.99 &	&	63,364.42	\\
120 & 65,172,721 & 1,001,821 & 1.79  & 7.01 &	&	85,984.84	\\
150 & 81,887,368 & 1,063,932 & 1.57  & 7.02  &	&	138,013.92	\\
180 & 98,513,201 & 1,121,039 & 1.40  & 7.02  &	&	189,760.54	\\
210 & 116,210,011 & 1,175,458 & 1.24   & 7.01 &	&	210,275.28	\\
240 & 133,219,590 & 1,218,175 & 0.73  & 6.98 &	&	166,183.94	\\
270 & 154,182,872 & 1,275,119 & 1.67  & 6.96 &	&	209,492.52	\\
300 & 170,933,996 & 1,305,333 & 0.40  & 6.96 &	&	247,981.78	\\
\bottomrule
\end{tabular}
\caption{Comparison of VR systems with different training data volume, showing the number of days the source data comprises, the number of user sessions, seed products per model, proportion of sessions contributing new products (SNP), the average length of the sessions and the CPU seconds needed to compute the respective model.}\label{w2v:table}
\end{table}
We relate both a model's performance and the effectiveness of data invested to data volume, using a number of KPIs:
\begin{enumerate}
\item \textbf{Number of products (\#products)}: The number of different product types our model contains recommendations for in the sense of coverage.
\item \textbf{Conversion Rate (CR)} is computed as outlined in Sec.~\ref{calc_conversion_rate}. 
\item \textbf{Revenue (rev)} is an approximation based on the number of products and the conversion rate, assuming constant values for visits per product and price of a product. We consider revenue the actual value generated by the algorithm.
\item \textbf{Average revenue per session} assigns each session a share of the overall revenue, following the simplified assumption that each session contributes equally to the overall revenue. 
\item \textbf{Sessions with new products (SNP)} relates the number of sessions to the number of new products, indicating which percentage of the sessions used for the model contains information on products that were not part of the next smaller model.
\end{enumerate}
\begin{figure}
\centering
\includegraphics[scale=1.0]{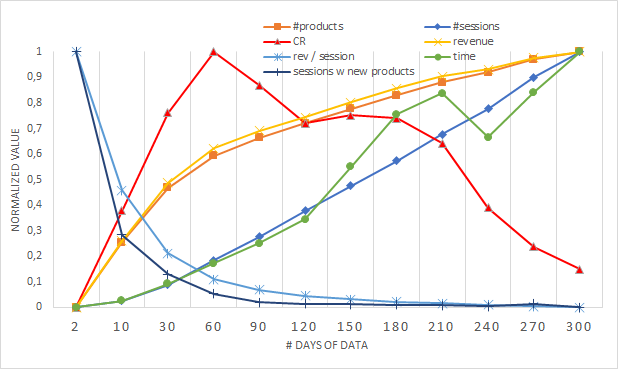}\caption{Different KPIs (feature-scaled between 0 and 1) related to model data volume.}
\label{w2v:curves}
\end{figure}
Table \ref{w2v:table} summarizes the basic results of this experiment, along with the proportion of added sessions containing new products and the computing time. As expected, the processing time rises with the amount of data, while the number of sessions contributing new seed products decreases.

Figure \ref{w2v:curves} shows how the KPIs relate to different models. The results give us a first indicator on both, how much data we actually need, and what kind of data is valueable: 
The CR (red line with triangles) shows an interesting trend when related to data volume. The CR peaks for the comparatively data-lean model comprising 60 days of data (20\% of the maximum). One reason for this is that more products (orange squares) come with a larger long-tail of rarely clicked products, for which the VR computes less relevant alternatives. While the revenue generated by the algorithm (yellow crosses) grows along the same trajectory as the number of products the model contains information for, it remains unclear whether the decrease in CR is inevitable or preventable by using the right data.

Most interestingly, the overall revenue grows more slowly at a model of intermediate size (120 days), where the conversion rate reaches a plateau before falling rapidly. Beyond this point, conversion rate drops far more rapidly than the number of new products rises when adding more data. This is further backed looking at the revenue per session. Naturally, more sessions mean less impact for each single session, and the revenue per session drops for larger models. At the same time, the number of new products that can be added to the model drops in the same fashion (cf. the two blue curves). This is in line with assumptions about standard learning processes: the curve for new products and value per session is equally distributed as the error rate in standard learning curves, which drops rapidly with more data initially, but then converges if more data is added.

While most KPIs converge along the learning curve, computing time increases almost linearly with data volume. An uninformed increase of the input data set will thus invest more and more time and data, and lose efficiency in terms of CR and value per data point. 

Our graph can give a first indication about which sessions are actually informative to the model, namely the ones containing new products.  Starting from this evidence on unequal information contribution of data points, we will elaborate on identification criteria of valuable sessions in the next section.

%% file: experiments.tex
\section{Data Value: \textbf{Which data} do we need?}\label{experiments}
We now outline how to define valueable data points. After explaining some assumptions and prerequisits (Section \ref{minimalism:assumptions}), we give a theoretical account of data value computation (Section \ref{minimalism:datavalue}) which we then put into practice in a lab-scale experiment (Section \ref{dsa:exp_setup}).

\subsection{Assumptions and Prerequisites}\label{minimalism:assumptions}
We compute the value of data based on three \textbf{assumptions}:

\begin{enumerate}

\item \textbf{The value of data must be computed at data point level.}
Frequently, the monetary value of data is computed as an aggregated value of a whole data source \cite{DBLP:journals/corr/abs-1811-04665}. For example, each data source could be assigned a value that sums up the revenue of all use cases using its data. We think that this kind of approach is too simplistic to implement data minimalism strategies. In our opinion, we need to consider the business value of individual data points to maintain the quality of data-driven decisions using as few data as possible. As a simple example, consider a customer interaction in which an unknown customer just clicks on the entry page of a shop and leaves again. This interaction does not contribute valuable information. However, if the very same customer clicks on an offer and a recommended article and ends up purchasing it -- that interaction is extremely valuable for computing customer preferences, better recommendations and many other customer-related decisions. 
Usually, a data source cannot be omitted as a whole, but in many cases, a large proportion of data points only carry redundant information and can be ignored when computing models or analyses. 

\item \textbf{The value of data exclusively depends on its contributions to decision quality.}
Data creates value by decision enablement. In order to compute the value of individual data points, we need to evaluate each data point's information contribution to a decision. Relating a data point's informational contribution to monetary value requires a quantifiable KPI that measures the decision's quality and, if possible, monetary profit. Ideally, the KPI not only detects beneficial data to the system at hand, but also reflects the worsening of a decision with negative values (i.e less profit). Value-neutral data points may fail to give an indication with the abovementioned KPI, since the measured profit does not necessarily vary with the output.

\item \textbf{Data-driven decisions can be automated.}
Knowing that this is currently not true, we make the bold assumption that in the near future, virtually all data-driven decisions will be automated. While we strongly believe that humans will always be involved in all core decisions made in a corporation, automated decisions surpass human decisions in orders of magnitude, independently of the decisions' relevance. We also believe data minimalism will mainly matter for automation, and hardly ever for data-driven decisions on a scale that humans can oversee.
\end{enumerate}

To accommodate those points, systems need to fulfill three \textbf{prerequisites}:
\begin{enumerate}
\item \textbf{Stability:} Machine learning systems are mostly non-deterministic: two models trained on the same data can yield different results in identical contexts. Those instabilities increase with the algorithm's complexity and the data volume. Such undesired instabilities are side-effects of desired features, because processing large amounts of data requires algorithms to sample. This often leads to random decisions between equally good options, which guarantee a stable average performance. 
In order to measure a data point's contribution to the overall performance, it is necessary to separate the data information's influence from algorithmically predisposed instabilities. We will show approaches to do so in the next section.
\item \textbf{Qualitative output evaluation:} Computing both system stability and data point impact requires to measure output consistency. Quantitative KPIs do not cover this entirely, because a system can reach the same performance with a different output. For e.g. similarity rankings, one can generate a stable set of testseeds and approximate the actual output consistency by tracking changes in the rankings for a sufficiently large test set.
\item \textbf{Quantitative performance indicators:} Quantitative KPIs are essential to compute a data point's value. While a data point that does not impact the system output is always meritless, data points that do change the output can have a positive value, a negative value (corrupting system performance) or no value at all (changing the output but not the performance). Our main KPI is revenue, along with conversion rate and computing time as additional indicators. Other systems and contexts might require different KPIs for optimization.
\end{enumerate}

\subsection{The value of a data point}\label{minimalism:datavalue}
\begin{figure*}
\includegraphics[scale=.5]{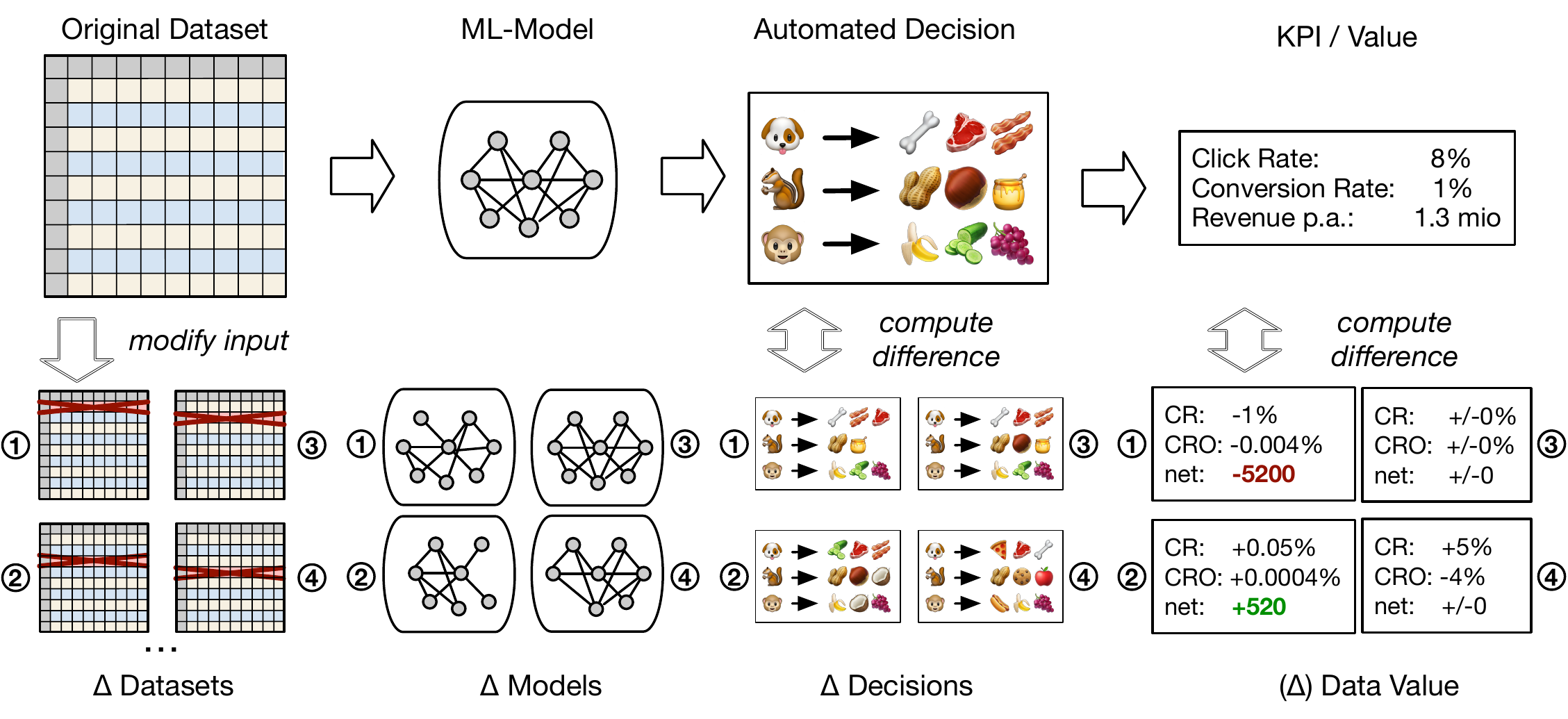}
\caption{Computing data value via sensitivity analysis}
\label{datavalue:img}
\end{figure*}

In order to build a data-driven system with a minimal data basis that achieves maximal positive impact, we need to identify valuable data points and distinguish them from redundant or even toxic ones. Both redundant and misleading information are always considered relative to a model trained on data points previously added to the system. Our final goal is to predict which new data points are worth considering and thus compensate costs with their value.
Figure~\ref{datavalue:img} shows the 3-step procedure we propose to estimate the value of a data point with respect to a model.

First, we undertake a sensitivity analysis designed as leave-one-out experiment (cf. \cite{DBLP:journals/corr/abs-1807-07404}). We test an otherwise unchanged data-driven system with different input data: the full system uses the whole dataset. Each test systems' input is the original dataset with the tested data point omitted (a $\Delta$ dataset). 
The system with a $\Delta$ dataset as input produces a $\Delta$~model. Note that we do not evaluate differences between individual $\Delta$~models, but rather compare each $\Delta$~model with the original model (differing in exactly one data point as input).

We repeat this procedure for a sample of data points which is sufficiently large to produce significant results while maintaining computational tractability. 
While such a random sample of test data points can reflect the proportion of data without influence, an automated predictive analysis of data value might require a more balanced or even exhaustive set of data points (which was neither computationally tractable nor the current objective in our case).

In a second step, we  build an end-to-end test setup in order to compare the resulting  $\Delta$~models' decisions with those of the original model to determine their overall influence on the system's output. The nature of those decisions certainly depends on the system in place; in our illustrative example, we assume a classification task resembling a recommender engine (assigning foods to animals). Differences of two models can show in changes of foods displayed, or in changes of their ranking.  After this intermediate step, we can already filter out data without value. Data points that do not promote changes in the system's behavior can be safely omitted, because they cannot change the final KPIs either.




We finally derive the monetary value of the data point by measuring the (positive or negative) impact of its omission on the system's KPIs.  To accomplish this, we map informational change in the model to informational change that matters for the output: we translate the changes in the system's decisions to changes in the system's quality. In our case, quality is measured preferably in revenue, but conversion rate and computing time also serve as valid measures. Here four constellations can arise: 
\begin{enumerate}
\item Output does not change: If the output does not change in the first place, there is no change in the KPIs.
\item Output changes, KPIs do not change: Some output variances might result from choices between equally good options and thus do not impact performance. Data points giving rise to such qualitatively but not quantitatively measurable changes do not have a value either.
\item Output changes, KPIs rise: The omitted data point made a difference, but the system performs better without it. This data point has a negative value (and is a toxic data point).
 \item Output changes, KPIs fall: The omitted data point made a difference, and the system performs worse without it. This data point has a positive value.
\end{enumerate}
The actual value of the data point depends on the difference it makes within the system. Of course, the same data point has fluctuating values at different points in time, in different systems, and is dependent on the constellations of all other data points. 

In order to estimate the value of a data point, we compute the conversion rates of the $\Delta$ models and of the baseline model as described in Section~\ref{recos}. We then calculate the relative change in CR. Assuming CR to be proportional to revenue, we project the value of the data point as the relative change of the conversion rate multiplied by the revenue that corresponds to the baseline model.

In theory, this kind of analysis can be done with any data-driven system and setup. In order to compute the actual data value, however, we require stability as a prerequisite. Many systems are deterministic and thus stable by nature (for an example of deterministic association analysis, see Section~\ref{datavalue:det_sys_association_analysis}). Conversely, many machine learning models, in particular neural nets, are often instable and can produce different outputs when trained with the same input. Because we want to separate such algorithmic instabilities from informationally caused differences, we assume a stable system. Note that a subset of invaluable data can also be identified in instable systems: If the system shows constant KPIs for identical input, data points that do not change the KPIs can be safely assumed to have no value.



\subsection{Data Value in the Lab}
\label{dsa:exp_setup}
We apply our theoretical considerations to two lab-scale recommender systems.
The sampled data we use as input consists of 1,112,410 user sessions (2 days) containing 607,049 distinct products.  First, we analyze the impact of a single session on the ranking of the recommendations computed with the co-occurence recommender (COR, cf. Section \ref{datavalue:det_sys_association_analysis}) and the vector recommender (VR, cf. Section \ref{w2v_intro}). In a second step, we compute the value of single sessions for the VR.



\subsubsection{Impact of single sessions on the ranking of recommendations}
We
define the \emph{output change} of the recommender system as a change in the ranking of the top~5 recommendations for a product (either as a change in order, or as a change in recommended products), and analyze the two recommender systems.
\subsubsection*{COR --} Because the COR training procedure is computationally inexpensive, we iterate over each session, leave it out of the co-occurrence matrix and calculate the model's recommendation changes. Thus we determine the impact scores for each input data point.  37\% of the user sessions do not have any impact on the top~5 recommendations.
\subsubsection*{VR --} Keeping the calculation of VR $\Delta$~models computationally tractable requires a restriction of analyses to a sub-sample of left-out sessions. We create 500 $\Delta$~datasets by randomly leaving out a single session and compute recommendations with the VR algorithm. We consider the top~5 recommendations of 268,173 seed products, omitting products occurring less than 5 times. We choose the same setup as described in Section~\ref{w2v_intro}, except for the number of threads. As discussed in a previous research paper \cite{DBLP:journals/corr/abs-1807-07404}, thread-concurrency causes unpredictable instabilities. To comply with our prerequisite of stability, we compute the model training single-threaded.
As a result, 12 sessions ($2.4\%$) do not have any influence on the output, showing, as expected, a higher sensitivity in the machine learning algorithm than in the deterministic COR.

\subsubsection{Impact of single sessions on conversion rate}



\begin{figure}
\includegraphics[scale=0.35]{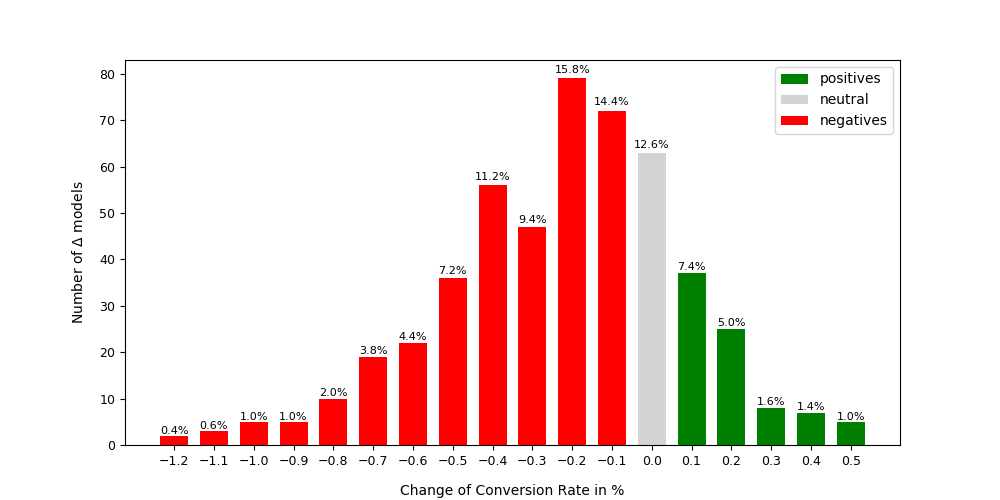}
\caption{Relative change of conversion rate for the vector recommender.}
\label{w2value}
\end{figure}

We quantify the impact of the left-out session by calculating the change in conversion rate (cf. Sec.~\ref{recos}) as described in Section~\ref{minimalism:datavalue}.
For the COR, the conversion rate only changes marginally ($0.002\%$, reaching up to $0.077\%$), which we thus do not analyze any further. In contrast, using the VR algorithm, the relative CR change is comparatively large. 
Figure \ref{w2value} shows the frequency distribution of this quantity for the 500 $\Delta$~models. The x-axis shows the CR change relative to the baseline model. Each bar aggregates a bin of models with a CR change within a 0.1 range.  The y-axis represents the number of models within a bin. For instance the green bar on the far most right contains 1.0\% of all models (5 models) with an increased CR of $0.5\%$.
The baseline model as well as 62 models with virtually unchanged conversion rates ($\pm 0.05\%$) are combined and represented by the grey bar. All other models show either an increase (green) or a decrease (red) in CR. 

\vspace{-0.75cm}

On average, the CR changes by $\pm 0.289\%$, with a minimum and maximum value of $-1.233\%$ and $0.546\%$, respectively. Based on those results, we model the value of individual sessions: we attribute the changes in CR ($\Delta CR$) for each model to leaving out one specific session, and assume the difference in revenue, caused by a difference in conversion, to be the inverse session value (assuming constant values for site visits). We thus compute the value of a session as follows, with $revenue$ being the overall revenue generated by the VR recommender:
$$
value(session) = -1 * \Delta CR * revenue_{VR} 
$$
In our example, a CR increase by $0.5\%$ means the overall revenue increases by $0.005$ when leaving out that specific session. This session thus has a \emph{negative} value of $-0.005 * revenue_{VR}$.
Analogously, a conversion rate decrease of $1.2\%$ means the overall revenue decreases by $0.012$ when leaving out that specific session. Thus this session adds a \emph{positive} value of $ 0.012 * revenue_{VR} $.
For a hypothetic revenue of $100,000,000\,\$$ generated by the vector recommender VR, 
a value of 1,200,000\,\$ would be attributed to that session. For other KPIs (e.g. click-through rate), a non-monetary value can be computed in the same fashion.


%% file: conclusion.tex
\section{Conclusion}\label{conclusion}
We conclude this paper with a short summary of our main experiments and contributions, followed by an outlook on future steps towards a full account of data value and fully implemented data minimalism.

\subsection{Summary}
We introduced data minimalism as a desirable paradigm, even if there is no actual lack of data. To support those claims, we showed how much data serves as input for corporate-scale recommender systems without contributing to decision quality  (up to 23\% in our case). 
As an essential tool to become more data-efficient while maintaining performance, we motivated why and how to apply sensitivity analysis as a means to compute the value of individual data points. We then applied this method and showed first approaches that can classify valuable from redundant data in a predictive setting.
Our main contributions are  a proposal on context-sensitive computation of data value, and a strategy to develop data-minimalistic systems. 

\subsection{Future Work}
Both the computation of data value and the implementation of data\hyp{}minimalistic systems are complex. Our work relies on several assumptions about automated decision systems which might be hard to comply with: Primarily, stability is uncommon in sophisticated AI systems, in particular neural nets. Further, not every automated decision system can be easily evaluated, let alone be monetarily priced.
With both of those shortcomings, we only delivered a laboratory\hyp{}scale implementation of a paradigm that is designed for corporate\hyp{}scale business. 

In order to scale and generalize our method to arbitrary automated systems, more research on predicting  data value is necessary. Most importantly, qualifying features that distinguish a data point as useful, redundant or toxic with respect to a given system could significantly contribute to more transparent AI algorithms, optimized output quality, higher system predictability and sustainable data minimalism.
We believe that with such progress, any system can become data\hyp{}minimal, which might turn out to be a vital AI strategy given the explosion of available data we will face in the upcoming years. We plan to move forward towards better explainable and better measurable systems - and thus reduce data consumption for the social, economical and ecological good.